\documentclass[letterpaper, 10 pt, conference]{ieeeconf}  

\IEEEoverridecommandlockouts                              

\usepackage{float}
\usepackage{amsmath,amssymb,amsfonts}
\usepackage{graphicx}
\usepackage{textcomp}
\usepackage{cuted}
\usepackage{subcaption}
\usepackage{comment}
\usepackage{tabularx}
\usepackage{booktabs} 
\usepackage{multirow}
\usepackage{url}
\usepackage{dcolumn}
\usepackage{multicol,lipsum}
\usepackage{mathtools, nccmath}
\usepackage{cuted}
\usepackage{flushend}
\usepackage{amssymb}
\usepackage{tabularx, lipsum}
\usepackage{soul}
\usepackage{booktabs}
\usepackage{collect}
\usepackage{verbatim}
\usepackage{booktabs}
\usepackage{multirow}
\usepackage{caption}
\usepackage{amsmath}
\usepackage{lipsum}
\usepackage[noend]{algpseudocode}
\usepackage{algorithm,algorithmicx}
\usepackage{kotex}
\usepackage{dsfont}
\usepackage{subcaption}
\usepackage{graphics}
\usepackage[normalem]{ulem}
\usepackage{cite}
\useunder{\uline}{\ul}{}
\usepackage{fancyhdr} 
\fancypagestyle{firstpageheader}{
    \fancyhf{}  
    \fancyhead[C]{\textit{This work has been accepted for publication at the \\ IEEE/RSJ International Conference on Intelligent Robots and Systems (IROS), 2025, \copyright~IEEE}}
    \vspace{8pt}
}

\title{\vspace{1.5em}
\LARGE \bf
Decision Transformer-Based Drone Trajectory Planning with Dynamic Safety–Efficiency Trade-Offs
}

\author{Chang-Hun Ji$^{1}$, SiWoon Song$^{2}$, Youn-Hee Han$^{1, *}$, SungTae Moon$^{2, *}$
\thanks{$^{1}$Future Convergence Engineering, Korea University of Technology and Education, Cheonan 31253, South Korea
        {\tt\small koir5660@koreatech.ac.kr, yhhan@koreatech.ac.kr}}%
\thanks{$^{2}$Department of Intelligent System $\&$ Robotics, Chungbuk National University, Cheongju, 28644, South Korea
        {\tt\small stmoon@cbnu.ac.kr, siwoons7319@cbnu.ac.kr}}%
\thanks{*Corresponding Authors. 
{\tt\small yhhan@koreatech.ac.kr, stmoon@cbnu.ac.kr}}%
}

\begin{document}
\maketitle
\thispagestyle{firstpageheader}  

\begin{abstract}

A drone trajectory planner should be able to dynamically adjust the safety–efficiency trade-off according to varying mission requirements in unknown environments.
Although traditional polynomial-based planners offer computational efficiency and smooth trajectory generation, they require expert knowledge to tune multiple parameters to adjust this trade-off.
Moreover, even with careful tuning, the resulting adjustment may fail to achieve the desired trade-off.
Similarly, although reinforcement learning-based planners are adaptable in unknown environments, they do not explicitly address the safety-efficiency trade-off.
To overcome this limitation, we introduce a Decision Transformer–based trajectory planner that leverages a single parameter, Return-to-Go (RTG), as a \emph{temperature parameter} to dynamically adjust the safety–efficiency trade-off.
In our framework, since RTG intuitively measures the safety and efficiency of a trajectory, RTG tuning does not require expert knowledge.
We validate our approach using Gazebo simulations in both structured grid and unstructured random environments.
The experimental results demonstrate that our planner can dynamically adjust the safety–efficiency trade-off by simply tuning the RTG parameter.
Furthermore, our planner outperforms existing baseline methods across various RTG settings, generating safer trajectories when tuned for safety and more efficient trajectories when tuned for efficiency.
Real-world experiments further confirm the reliability and practicality of our proposed planner.

\end{abstract}


\section{INTRODUCTION}
A drone trajectory planner should dynamically adjust the trade-off between safety and efficiency based on specific mission requirements \cite{umer2025cognitive}.
However, traditional polynomial-based planners, which are widely used due to their computational efficiency and smooth trajectory generation, require substantial expert knowledge for tuning multiple parameters to effectively adjust this trade-off \cite{arshad2023quadrotor}.
Even after careful parameters tuning, polynomial-based planners tend to prioritize efficiency, resulting in trajectories that may not dynamically adapt to changes in the safety–efficiency trade-off \cite{zhou2019robust, penicka2022learning}.
Similarly, although reinforcement learning (RL)-based planners have gained popularity due to their adaptability in unknown and complex environments, RL approaches do not explicitly account for the safety–efficiency trade-off \cite{yu2024pathrl, rocha2024enhancing}.
Consequently, adjusting this trade-off using RL-based planners necessitates iterative reward function design and extensive retraining processes, resulting in significant computational and time costs \cite{devlin2012dynamic}.

To overcome these limitations, we propose a planner based on Decision Transformer (DT) \cite{chen2021decision}. 
DT offers a new perspective on control: rather than maximizing a reward function, DT allows a policy to be conditioned on a desired return-to-go (RTG), which enables the policy to generate actions to achieve that return \cite{zheng2022online, wu2023elastic, xie2023future}. 
Consequently, by varying the conditioning RTG value, a single trained DT model can produce diverse behaviors.
By applying this approach, we propose a drone trajectory planner that uses DT to flexibly adjust the trade-off between safety and efficiency.
Our insight is that we can leverage RTG as a \emph{temperature parameter}, which can be tuned to dynamically adjust safety and efficiency within a single trained model.

\begin{figure}[t!]
    \centering
    \includegraphics[width=\linewidth]{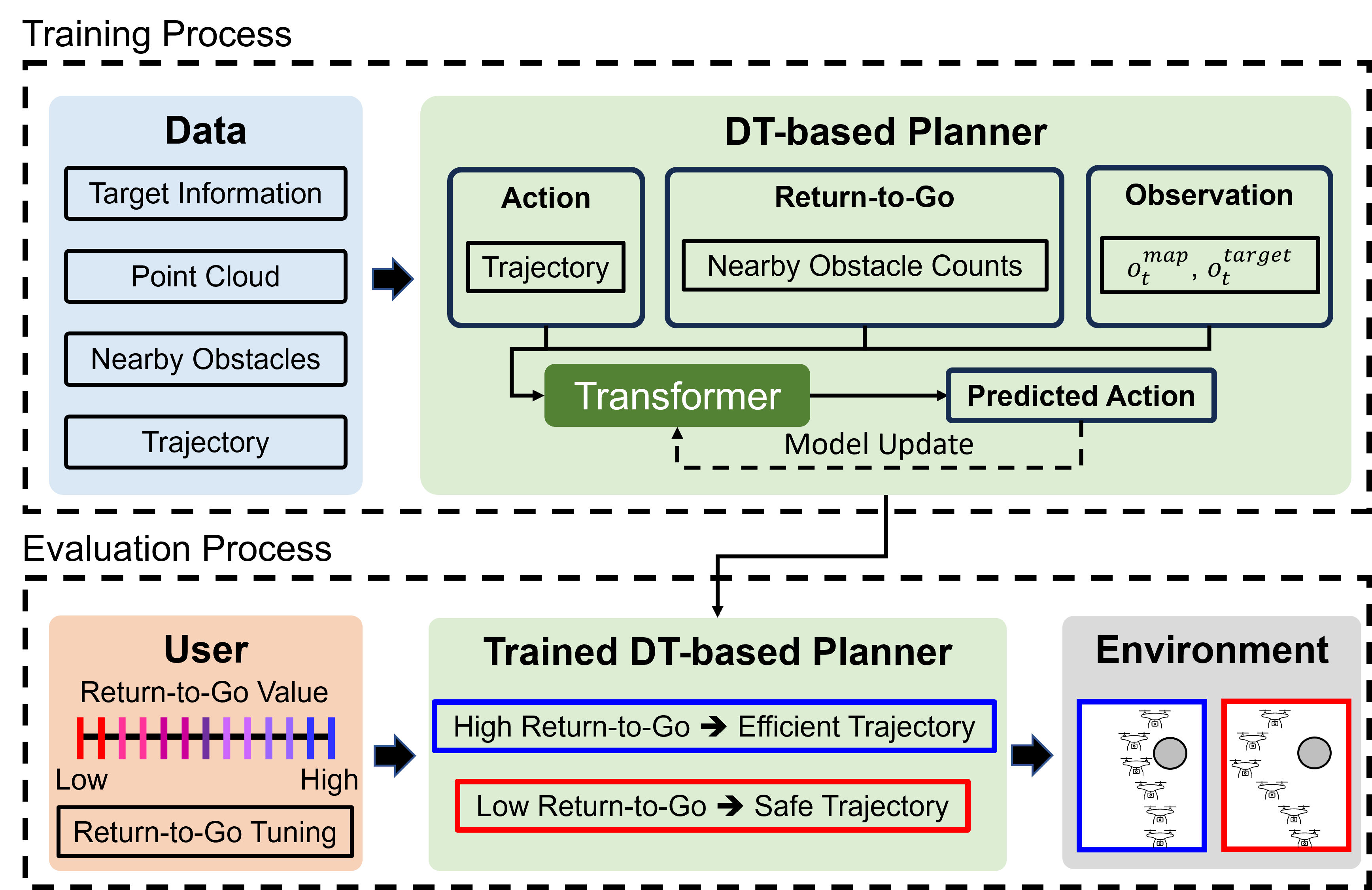}
    \caption{Overview of our Decision Transformer (DT)-based planner framework. After training, users can simply adjust the Return-to-Go to dynamically adjust the safety–efficiency trade-off without retraining or separate models.}
    \label{fig:overview}
\end{figure}

In our framework, the reward function intuitively measures efficiency and safety of a trajectory.
Therefore, given a specific RTG, our DT‑based planner can prioritize either trajectory safety or efficiency to satisfy the desired RTG.
This enables users to dynamically adjust the safety-efficiency trade-off by simply tuning the RTG without additional models or retraining.

However, conventional DT methods predict actions satisfying an RTG computed over the entire remaining episode. 
This makes it difficult for users to specify an appropriate RTG for obtaining desired behaviors in unknown environments.
To address this issue, we introduce an $N$-step RTG, which limits the prediction horizon to a short, predictable window of $N$ future steps.
The intuitive design of the reward function and the proposed $N$-step RTG enable users to effectively tune the desired RTG without requiring expert knowledge.
Moreover, to ensure training‑data diversity by balancing both safety and efficiency, we have developed a medial axis-based planner that explicitly computes safe reference trajectories.
As shown in Fig. \ref{fig:overview}, our approach allows users to intuitively adjust the safety–efficiency trade-off by simply tuning the RTG.

Experimental results in various Gazebo simulations demonstrate that our approach can flexibly adjust the trade-off between safety and efficiency with a single parameter, RTG.
Furthermore, we observe that our planner outperforms existing baseline methods across various RTG settings, producing safer trajectories when RTG is tuned for safety and more efficient trajectories when tuned for efficiency.
Finally, we validate the practical applicability of our method by deploying it in real-world scenarios, demonstrating its capability to dynamically control the trade-off between safety and efficiency in trajectory generation.

The key contributions of this paper are as follows.
\begin{itemize}
    \item We propose a DT-based planner for autonomous drone flight in complex and unknown environments.
    \item We introduce RTG as a temperature parameter to dynamically adjust safety-efficiency trade-off within a single trained model, eliminating the need for additional training or reward adjustments.
    \item The simulation experiments demonstrate that our approach effectively modulates safety and efficiency, outperforming existing baseline methods according to the intended tuning of the RTG.
    \item Real-world flight tests validate the practical effectiveness and feasibility of our proposed algorithm for actual drone operations.
\end{itemize}


\section{RELATED WORK}
\subsection{Decision Transformer}
DT is a novel reinforcement learning framework that formulates sequential decision-making as a conditional sequence modeling task, leveraging transformer architectures originally developed for natural language processing tasks \cite{chen2021decision, zheng2022online, wu2023elastic, xie2023future}. 
Unlike traditional reinforcement learning methods that directly optimize policies by maximizing expected returns, DT follows an offline reinforcement learning paradigm, where the model is trained on pre-collected offline datasets without further online exploration.
Specifically, DT generates actions by conditioning on a specified future cumulative reward, known as RTG, enabling flexible and dynamic behavior adjustments based on offline expert data.

RTG at timestep $t$, denoted as $\text{RTG}_t$, is defined as the cumulative sum of immediate rewards from timestep $t$ through to the final timestep $T$ of the episode, mathematically expressed as $\text{RTG}_t = \sum_{k=t}^{T} r_k$, where $r_k$ represents the immediate reward at timestep $t$.
During the training phase, DT is fed sequences of past observations, actions taken, and received rewards, paired with their corresponding RTGs. 
The transformer architecture \cite{VaswaniSPUJGKP17}, leveraging self-attention mechanisms, processes these sequences to predict actions most likely to achieve the provided RTG. 
Specifically, the DT model learns to associate different RTG values with appropriate actions by minimizing prediction errors through gradient-based optimization. 
As a result, the trained model can dynamically adjust its behavior at inference time simply by changing the RTG parameter, enabling flexible trade-offs between competing objectives such as safety and efficiency without retraining.

\subsection{Polynomial-Based Planners}
Polynomial-based planners compute flight paths by optimizing predefined objective functions. 
Recent works have proposed planners generating polynomial segments within safe flight corridors \cite{9310337} or spherical corridors with receding-horizon planning for high speed flight in complex environments \cite{ren2022bubble}. 
TGK-Planner efficiently produces dynamically feasible trajectories for aggressive flights \cite{ye2020tgk}, while several other approaches utilize B-splines to ensure smoothness and dynamic feasibility \cite{zhou2022swarm, xu2023vision, zhou2021raptor}.

Polynomial-based planners offer computational efficiency and smooth trajectory generation. 
However, tuning the parameters of polynomial-based planners typically requires substantial expert knowledge, making it difficult for users to effectively manage parameters related to the safety-efficiency trade-off \cite{arshad2023quadrotor}.
Furthermore, since polynomial-based planners inherently tend to prioritize efficiency, even meticulous parameter tuning may result in trajectories that do not significantly vary in terms of the safety-efficiency trade-off \cite{zhou2019robust, penicka2022learning}.

\subsection{Reinforcement Learning-Based Planners}
RL-based planners have been widely studied due to their adaptability and generalization capabilities in unknown environments.
Low-level RL-based planners that directly control drone actions enable real-time autonomous operation and navigation \cite{song2021autonomous, hodge2021deep} and provide effective 3D path planning by integrating optimization strategies \cite{yu2023reinforcement}. 
However, these direct control methods often suffer from instability and a lack of trajectory smoothness, particularly during extended flights \cite{wang2023efficient}.
Consequently, recent research actively explores high-level RL-based planners that generate stable and smooth trajectories suitable for long-term navigation tasks.

High-level RL-based planners include methods that determine control points for Bézier curves to generate polynomial trajectories \cite{yu2024pathrl}, as well as distillation-based techniques that produce optimal trajectories using only local map information rather than entire environment maps \cite{rocha2024enhancing}.
While RL-based planners provide superior adaptability and flexibility, they typically require retraining new models to adjust the safety-efficiency trade-off.
To address this limitation, we propose a DT‑based polynomial trajectory planner that dynamically adjusts the safety–efficiency trade‑off in a single model.

\section{Proposed Method}

\subsection{Markov Decision Process and Training}

In this section, we formulate the trajectory planning task in an unknown environment as a Markov Decision Process (MDP), allowing DT to dynamically adjust the safety-efficiency trade-off.

\subsubsection{Observation ($o_t$)}
The observation is composed of two parts, $o_t = 
\{o^{{map}}_t \in \mathbb{R}^{w \times h \times d}, o^{{target}}_t \in \mathbb{R}^{12}\}$. 
$o^{{map}}_t$ represents the drone’s perception of the environment at time $t$. 
It is a $w \times h \times d$ 3D matrix centered on the current position of the drone, where $w$, $h$, and $d$ denote the width, height, and depth of the 3D occupancy grid, respectively.
Each cell $(i, j, k)$ in this matrix contains a value of 1 (i.e., $o^{map}_{t}(i, j, k)=1$) if an obstacle is present and 0 (i.e., $o^{map}_{t}(i, j, k)=0$) if the space is free. 
This occupancy grid allows the model to understand the surrounding obstacles and navigate accordingly. 
In our framework, when a final target is provided for a given trajectory planning task, a local target $o^{{target}}_t$ is extracted at every timestep to facilitate local planning. 
The local target $o^{{target}}_t$ is a 12-dimensional vector representing the relative state of the target and the drone's motion information.
The components of $o^{{target}}_t$ are detailed in Tab. \ref{tab:trajectory_variables}.

\begin{table}[b!]
    \centering
    \caption{Description of $o^{{target}}_t$.}
    \renewcommand{\arraystretch}{1.2}
    \small 
    \setlength{\tabcolsep}{4pt} 
    \resizebox{\columnwidth}{!}{
    \begin{tabularx}{\columnwidth}{>{\centering\arraybackslash}p{2.0cm} X}
        \toprule
        \textbf{Element} & \textbf{Description} \\
        \midrule
        $p^{t}_{i} - p^{c}_{i}$ & The target's relative position to the drone in each direction ($i \in \{x,y,z\}$). \\
        $v^{c}_{i}$ & Current velocity of the drone in each direction ($i \in \{x,y,z\}$). \\
        $v^{t}_{i}$ & Target velocity in each direction ($i \in \{x,y,z\}$). \\
        $a^{c}_{i}$ & Current acceleration of the drone in each direction ($i \in \{x,y,z\}$). \\
        \bottomrule
    \end{tabularx}
    }
    \label{tab:trajectory_variables}
\end{table}

\subsubsection{Action ($a_t$)}
In our framework, the action is represented as a fifth-order polynomial trajectory, which defines the motion of a drone over a short time horizon.
Each action is parameterized by the coefficients of a fifth-order polynomial for each spatial axis, ensuring smooth and dynamically feasible local trajectories.
Formally, the position of the drone, $p(\tau_t)$ at relative time $\tau_t$ within each trajectory of timestep $t$, is given by:
\begin{equation}
    p(\tau_t) = a_0 + a_1\tau_t + a_2\tau_t^2 + a_3\tau_t^3 + a_4\tau_t^4 + a_5\tau_t^5,
\end{equation}
where $a_0$, $a_1$, and $a_2$ correspond to the current position, velocity, and half of the acceleration of the drone, respectively.
Since these lower-order terms ($a_0$, $a_1$, $a_2$) are directly determined by the current state, DT only needs to predict the higher-order coefficients ($a_3$, $a_4$, $a_5$) to shape the trajectory while maintaining continuity.
Furthermore, as the polynomial trajectory is defined independently for each spatial axis (x, y, and z), DT outputs a 9-dimensional vector representing the coefficients $[a_3^x, a_4^x, a_5^x, a_3^y, a_4^y, a_5^y, a_3^z, a_4^z, a_5^z]$ for all three axes. 
This compact action representation enables the model to generate dynamically feasible trajectories that smoothly adapt to drone and environmental constraints.

\begin{figure}[t!]
    \centering
        \begin{subfigure}{0.22\textwidth}
        \centering
        \includegraphics[width=\linewidth]{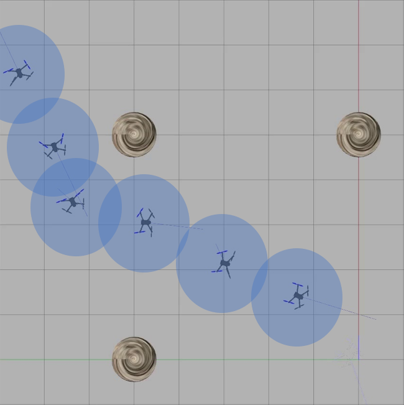}
        \caption{Safe trajectory}
        \label{fig:safe_reward}
    \end{subfigure}
    \hfill
    \begin{subfigure}{0.22\textwidth}
        \centering
        \includegraphics[width=\linewidth]{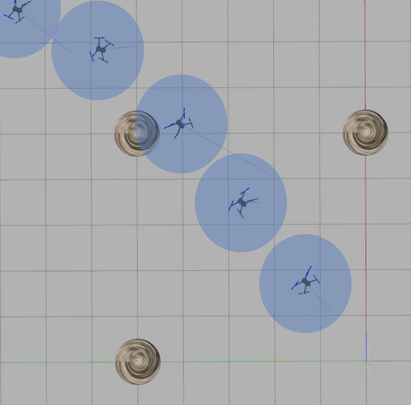}
        \caption{Efficient trajectory}
        \label{fig:efficiency_reward}
    \end{subfigure}
    \caption{An illustration of reward computation based on obstacle proximity. Fig. \ref{fig:safe_reward} shows a safe trajectory with no obstacles within the radius $\delta$ (blue circle), resulting in zero reward, while Fig. \ref{fig:efficiency_reward} shows an efficient trajectory with obstacles within $\delta$, yielding a positive reward.}
    \label{fig:reward}
\end{figure}

\subsubsection{Reward ($r_t$)} \label{reward_subsection}
The reward function is intuitively designed to indicate the trade-off between safety and efficiency rather than serving as an explicit optimization target. 
Specifically, at each timestep $t$, the reward $r_t$ is computed based on the distance $d_{i, j ,k}$ from the drone's current position to each voxel in the occupancy grid:
\begin{equation} d_{ijk} = \sqrt{\left(i - \frac{w}{2}\right)^2 + \left(j - \frac{h}{2}\right)^2 + \left(k - \frac{d}{2}\right)^2}. \end{equation}
Then, the reward $r_t$ is defined as:
\begin{equation} r_t = \sum_{i, j, k} o^{map}_{t}(i, j, k) \times \mathbb{I}(d_{ijk} < \delta), \label{eq:reward_function} \end{equation}
where $\delta$ represents the threshold distance for nearby obstacle calculation and $\mathbb{I}$ is the indicator function.
According to Eq. \eqref{eq:reward_function}, the reward at each timestep is based solely on the number of obstacles detected within a local region.

This seemingly simple measure effectively captures the safety-efficiency trade-off of a trajectory because the training data for our DT is generated from expert polynomial-based planners.
Therefore, the training data only contains trajectories that navigate toward the target while avoiding obstacles.
In these training datasets, many obstacles near the drone indicate that it is flying close to obstacles to reach the target efficiently. In contrast, a low obstacle count implies that the drone takes a broader, safer route.
As a result, the reward function in Eq. \eqref{eq:reward_function} naturally assigns lower rewards to conservative trajectories and higher rewards to more direct and efficient ones.
Fig. \ref{fig:reward} illustrates an example of this.

\subsubsection{Training} 
To achieve efficient training, we adopted a two-phase training scheme.
In the first phase, an autoencoder is trained to learn a latent vector representation of $o^{map}_t$.
Once the autoencoder training is completed, the encoder parameters are frozen to maintain the learned mapping from occupancy grids to compact latent vectors.
In the second phase, the DT models incorporating attention networks are trained.
During inference, the frozen encoder transforms $o^{map}_t$ into a latent vector, which is then concatenated with the embedding of $o^{target}_t$ before being fed into the DT model.

\subsection{Adjusting the Safety–Efficiency Trade-off via RTG}

DT selects actions to achieve the given RTG in future timesteps.
Therefore, we leverage RTG as a temperature parameter to adjust the trade-off between safety and efficiency in trajectories using a single trained model.
Specifically, since our reward function assigns higher rewards to trajectories closer to obstacles (efficient trajectories) and lower rewards to trajectories farther from obstacles (safe trajectories), at inference time, a higher RTG instructs DT to select actions that prioritize efficiency, resulting in trajectories closer to obstacles.
In contrast, a lower RTG encourages DT to prioritize safety, producing trajectories with larger safety margins.
This approach eliminates the need to train separate models for adjusting the safety-efficiency trade-off.
Furthermore, our reward function is intuitively designed to measure the safety and efficiency of trajectories, allowing users to effectively tune the RTG without requiring expert knowledge.
As a result, users can easily adjust drone trajectories to specific mission requirements.

\begin{figure}[t!]
    \centering
        \begin{subfigure}{0.23\textwidth}
        \centering
        \includegraphics[width=\linewidth]{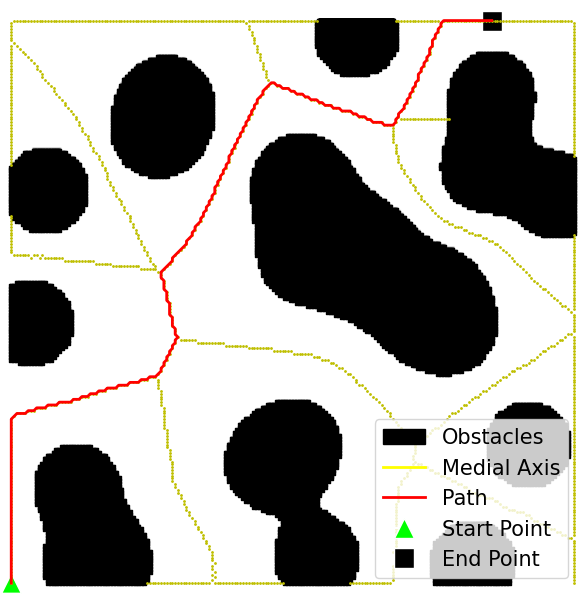}
        \caption{}
        \label{fig:only_medial_path}
    \end{subfigure}
    \hfill
    \begin{subfigure}{0.23\textwidth}
        \centering
        \includegraphics[width=\linewidth]{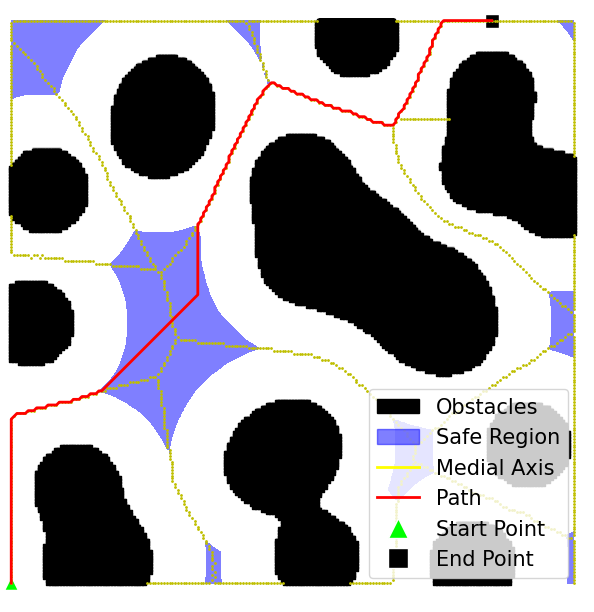}
        \caption{}
        \label{fig:medial_safety_region_path}
    \end{subfigure}
    \caption{Comparison of two approaches for safe trajectory references: Fig. \ref{fig:only_medial_path} uses only the medial axis, whereas Fig. \ref{fig:medial_safety_region_path} combines the medial axis with an obstacle-inflated safe region, producing safer and more efficient paths.}
    \label{fig:medial_axis_based_trajectory}
\end{figure}

\subsection{$N$-step RTG}
Conventional DT predicts actions based on the RTG, which represents cumulative rewards obtained until the end of an episode, introducing significant uncertainty in unknown environments.
For instance, a given RTG might represent a relatively low value in complex environments where higher rewards can be frequently obtained due to close proximity to obstacles.
At the same time, the same RTG could correspond to a relatively high value in simpler environments with sparse obstacles and lower per-step rewards.
This can hinder the effectiveness of RTG as a temperature parameter.

To address this issue, we propose an $N$-step RTG $G_t^{N}$, which truncates the return horizon to the next $N$ steps instead of the entire remaining episode. 
The $N$-step RTG is defined as:
\begin{equation}
G_t^{N} = \sum_{k = 0}^{N-1}r_{t+k},
\label{eq:n_step_rtg}
\end{equation}
where $N$ represents the number of steps for which DT can make reliable future predictions.
Therefore, training DT to select actions that achieve a desired RTG within a predictable $N$-step horizon enables effective and reliable behavior adjustments, even in unknown and dynamic environments.

\begin{figure}[t!]
    \centering
        \begin{subfigure}{0.2\textwidth}
        \centering
        \includegraphics[width=\linewidth]{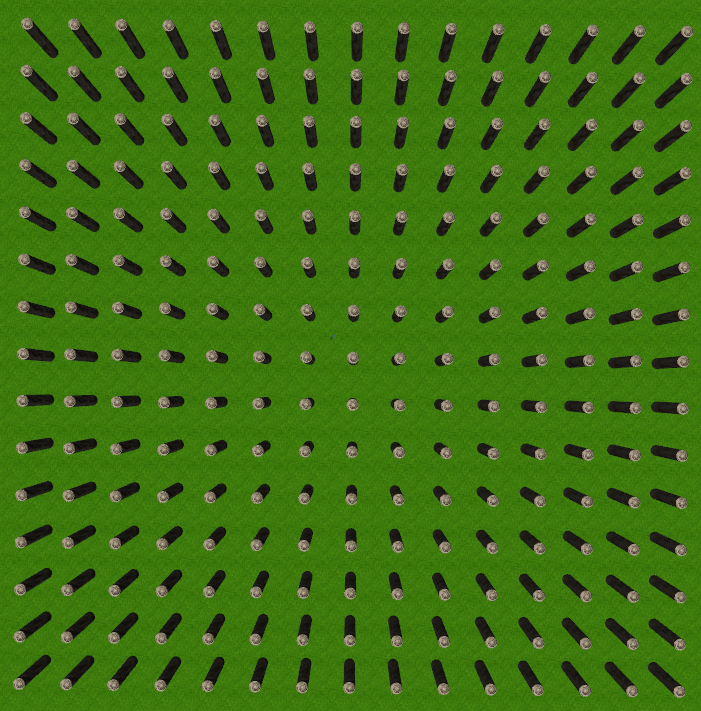}
        \caption{Grid map}
        \label{fig:grid_map}
    \end{subfigure}
    \hfill
    \begin{subfigure}{0.208\textwidth}
        \centering
        \includegraphics[width=\linewidth]{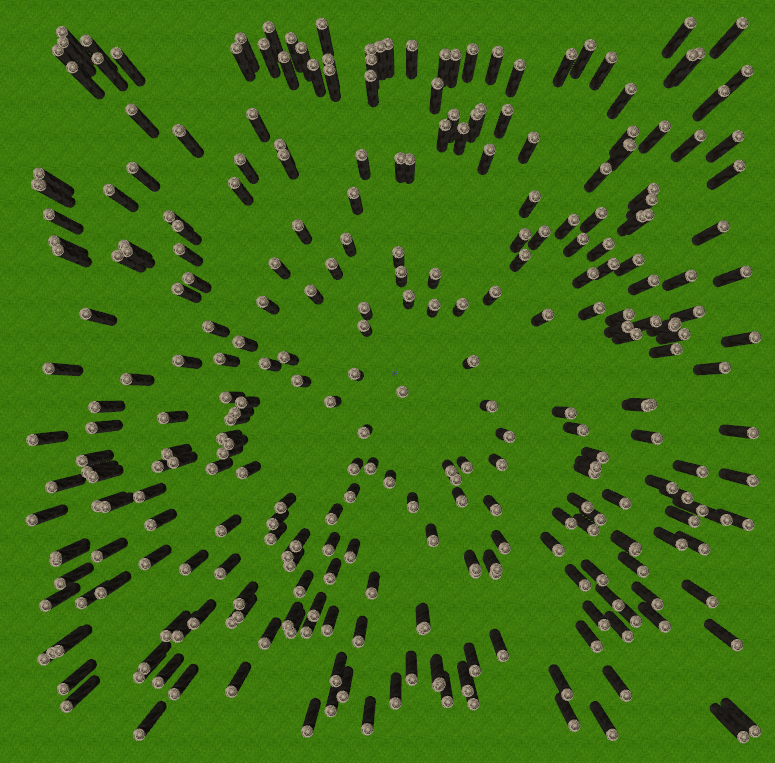}
        \caption{Random map}
        \label{fig:random_map}
    \end{subfigure}
    \caption{Examples of Gazebo simulation environments used in our experiments.}
    \label{fig:gazebo_env}
\end{figure}

\subsection{Expert Planners for Data Collection}
To enhance the training performance of DT in an offline RL setting, we construct a diverse dataset of expert drone trajectories across various environments.
To ensure high quality and diverse trajectories, we collect training data from multiple planning algorithms, each of which exhibits different strengths.

\subsubsection{EGO-Planner}
EGO-Planner \cite{zhou2020ego} is a state-of-the-art gradient-based local trajectory planning framework designed for real-time autonomous drone navigation. 
It focuses on efficiency, smoothness, and collision avoidance while ensuring dynamic feasibility.

\subsubsection{ViGO}
ViGO \cite{xu2023vision} is a trajectory planning framework designed to enable autonomous drones to navigate safely in complex, dynamic, and unknown environments. 
Unlike traditional planners that rely primarily on geometric constraints or pre-defined maps, ViGO integrates vision-based perception with optimization-based trajectory planning, allowing drones to adapt to new obstacles in real time.


\subsubsection{Medial Axis-Based Planner}
EGO-Planner and ViGO are widely recognized as state-of-the-art algorithms among polynomial-based planners.
However, polynomial-based planners primarily focus on efficiency and often fail to guarantee sufficient safety margins along the trajectory \cite{zhou2019robust, penicka2022learning}.
Therefore, we devise a medial axis-based planner to provide a safe trajectory reference.
 
Our method leverages the Generalized Voronoi Diagram \cite{wen2024g} and an obstacle inflation mechanism \cite{karlsson2023d}.
In our approach, both obstacles and the boundaries of $o^{map}_t$ are treated as obstacles to compute the medial axis of the free space, which yields a skeleton representation by identifying points equidistant from these edges and maximizing clearance.

To further enhance the smoothness and efficiency of the generated path while maintaining safety, we combine the medial axis with an obstacle-inflated safe region.
In this approach, obstacles are inflated by a predefined margin, and the remaining area is defined as a safe region for navigation, enabling the planner to produce smoother and more efficient paths.
We then construct a connectivity network along both the medial axis and the safe region, and apply a graph search algorithm (e.g., A*) to find an optimal path from the start to the target.
This path is subsequently converted into a smooth polynomial trajectory using a fifth-order polynomial parameterization.
Fig. \ref{fig:medial_axis_based_trajectory} illustrates a comparison between paths generated using only the medial axis and those that incorporate the safe region. 
As shown, our approach yields smoother and more efficient  trajectories while still maintaining safety.
Consequently, our medial axis-based planner provides safe expert trajectories for our dataset, prioritizing obstacle avoidance while maintaining a balance between safety and efficiency.

\begin{table}[b!]
    \centering
    \caption{The parameter settings of our planner training.}
    \label{table:learning_parameters}
    \footnotesize
    \begin{tabular}{p{5cm}p{2.5cm}}
        \toprule
        \textbf{Parameters}   & \textbf{Values} \\
        \midrule
        Learning rate of the autoencoder model & 0.001 \\ 
        Learning rate of the DT model & 0.0001 \\
        Number of heads in the DT model & 4 \\ 
        Weight decay for the DT model & 0.0001 \\
        Size of  $o_t^{map}$, \(w \times h \times d\) & \(100 \times 100 \times 10\) \\ 
        The threshold distance for reward, $\delta$ & 20 \\
        $N$-step return horizon, $N$ & 2 \\ 
        \bottomrule
    \end{tabular}
\end{table}

\section{Experiments}

\subsection{Experimental Setup}
We collected data for DT training and conducted simulation experiments using Gazebo simulations.
Specifically, expert trajectory data for training the DT were gathered from three planners (EGO-Planner, ViGO, and a medial axis-based planner), each running at a target velocity of 0.8 m/s with a replanning frequency of 2 Hz.
This diverse dataset enabled our planner to effectively learn both safe and efficient trajectories.
To ensure robust learning and comprehensive evaluation, we generated two distinct types of environments for training and simulation experiments.

\subsubsection{Grid Maps}
Structured obstacle layouts consist of obstacles arranged in a grid pattern.
The spacing between obstacles varies to create diverse grid maps.
In these maps, the safest route is typically along the medial axis of the free space, so a medial axis-based planner provides an ideal safety reference.

\subsubsection{Random Maps}
Random maps feature unstructured obstacle placements that force planners to handle novel and unpredictable scenarios.
These environments do not have one obvious optimal trajectory, which makes them more challenging and requires the model to generalize beyond the structured cases.
Fig. \ref{fig:gazebo_env} presents representative examples of the simulated environments.

\begin{figure}[t]
    \centering
        \begin{subfigure}{0.48\textwidth}
        \centering
        \includegraphics[width=\linewidth]{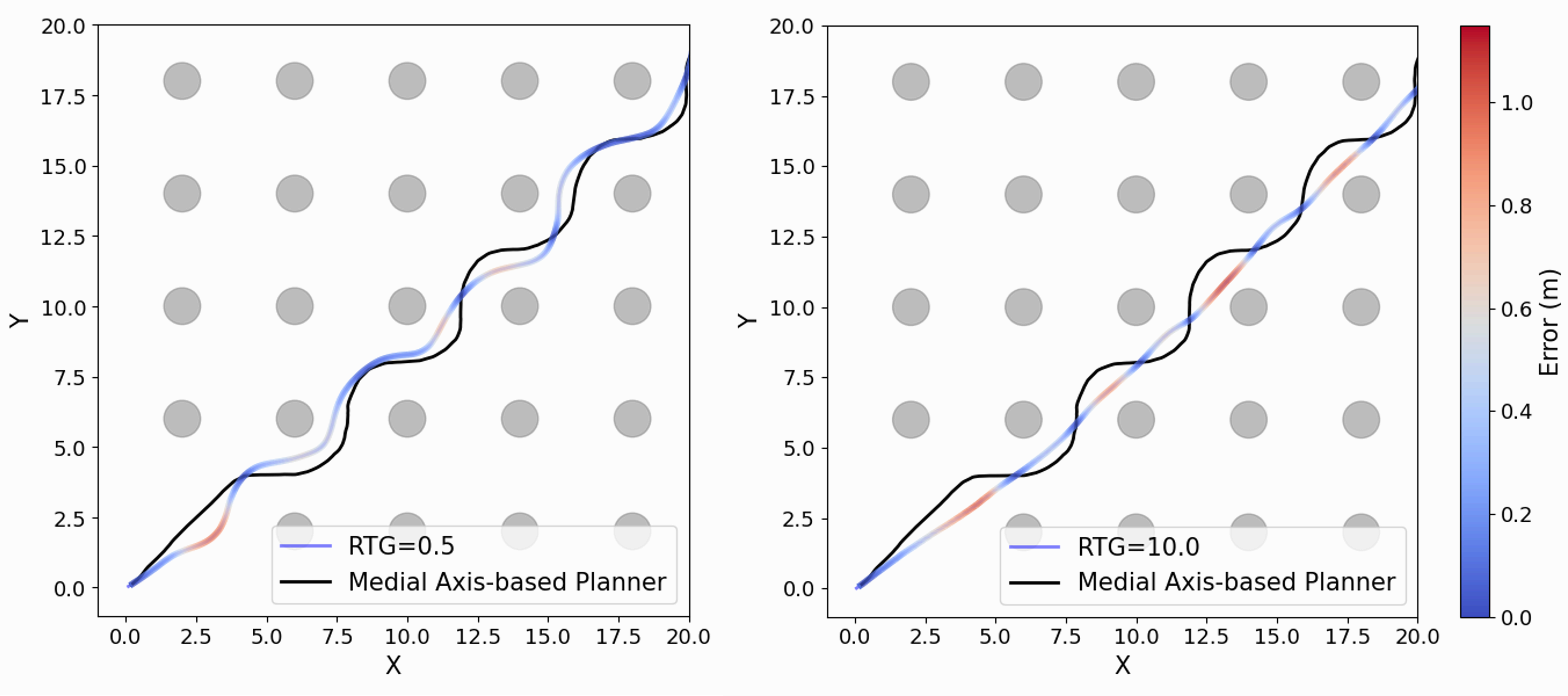}
        \caption{Grid map 4}
        \label{fig:grid_map_4_comparison}
    \end{subfigure}
    \hfill
    \begin{subfigure}{0.48\textwidth}
        \centering
        \includegraphics[width=\linewidth]{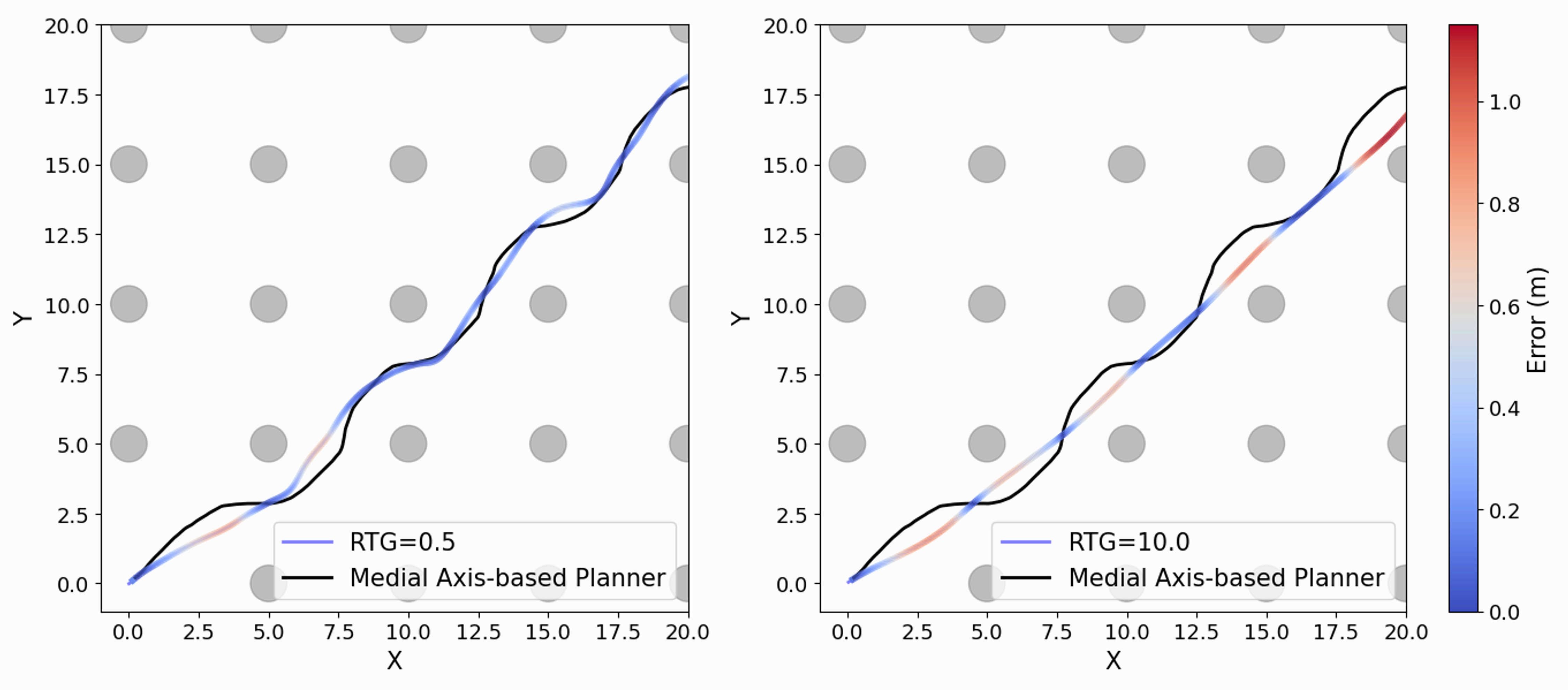}
        \caption{Grid map 5}
        \label{fig:grid_map_5_comparison}
    \end{subfigure}
    \caption{Comparison of trajectory generation and DTW error across varying RTG values in grid maps.}
    \label{fig:grid_map_comparison}
\end{figure}

\begin{table}[t]
    \centering
    \caption{Average performance of planners on grid maps.}
    \label{tab:grid_performance}
    \setlength{\tabcolsep}{3pt}
    \small
    \begin{tabular}{@{}lcccc@{}}   
        \toprule
        \textbf{Method}
        & \shortstack{Flight Dist\\(m)}
        & \shortstack{Avg Vel\\(m/s)}
        & \shortstack{\textbf{Flight Time}\\(s)}
        & \shortstack{\textbf{Safety}\\(DTW~Error)} \\  
        \midrule
        EGO-Planner                            & 31.24 & 0.71 & 43.44 & 0.46 \\
        ViGO                                   & 31.59 & 0.64 & 49.54 & 0.47 \\
        Medial-based                           & 34.05 & 0.39 & 84.97 & \textbf{0.00} \\
        Safe EGO                               & 31.03 & 0.71 & 43.09 & 0.44 \\
        \shortstack[l]{\textbf{Ours}\\[-2pt]\textbf{(RTG = 0.5)}}  & 33.72 & 0.67 & 49.41 & \underline{\textbf{0.32}} \\
        \shortstack[l]{\textbf{Ours}\\[-2pt]\textbf{(RTG = 10.0)}} & 32.16 & 0.92 & \underline{\textbf{37.32}} & 0.47 \\
        \bottomrule
    \end{tabular}
\end{table}

To collect a diverse dataset, we changed the map after every 100 trajectories.
For the grid maps, obstacle spacing was randomly set within a range of 2–4 m.
In the case of random maps, the number of obstacles varied randomly between 300 and 400.
We trained our proposed algorithm using this diverse dataset, collected from the simulation environments described above.
The detailed training hyperparameters are provided in Tab. \ref{table:learning_parameters}.

\begin{figure}[t]
    \centering
        \begin{subfigure}{0.23\textwidth}
        \centering
        \includegraphics[width=\linewidth]{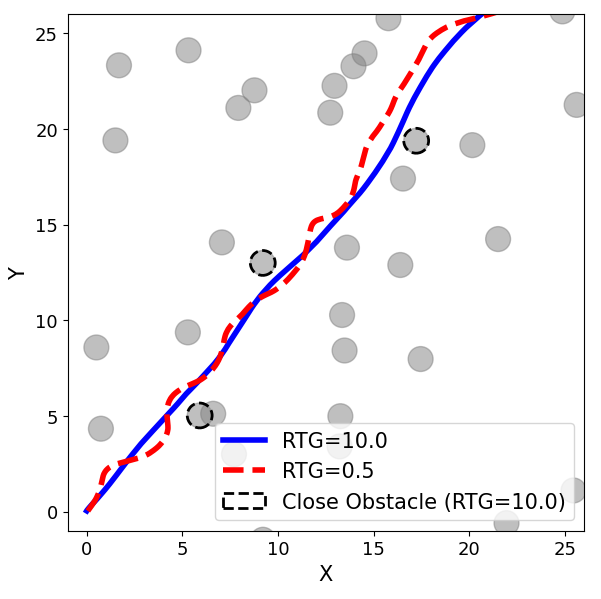}
        \label{fig:random_400}
    \end{subfigure}
    \hfill
    \begin{subfigure}{0.23\textwidth}
        \centering
        \includegraphics[width=\linewidth]{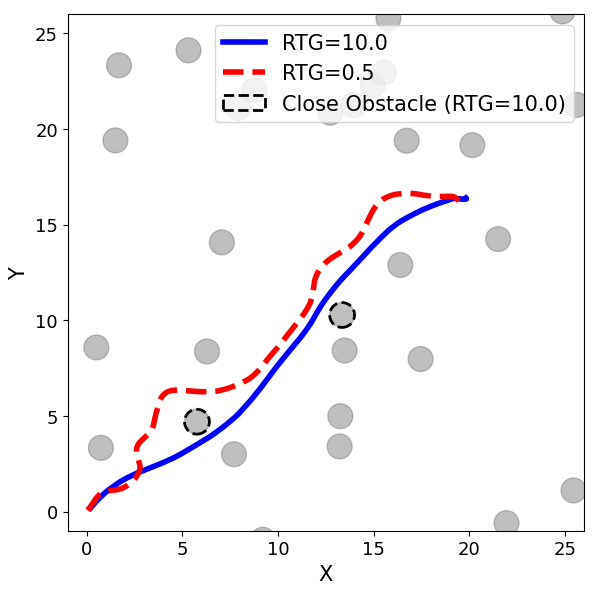}
        \label{fig:random_403}
    \end{subfigure}
    \caption{Comparison of trajectory generation across varying RTG values in random maps.}
    \label{fig:random_map_comparison}
\end{figure}

\subsection{Simulation Experiments}

In our simulation experiments, we evaluated our proposed algorithm on both grid and random map environments. 
We conducted experiments that included environments not encountered during training to test generalization further.

\subsubsection{Grid Map Experiments}
Our medial axis-based planner consistently produces the safest trajectory in grid map environments by maximizing obstacle clearance, serving as an optimal reference. 
Therefore, we quantify safety by computing the Dynamic Time Warping (DTW) error between the medial axis-based planner and the other trajectories \cite{gong2022motion}.
Concurrently, trajectory efficiency is defined as the ability to minimize flight distance and maximize average velocity to reach the endpoint.
Accordingly, we evaluate efficiency using flight distance (Flight Dist), average velocity (Avg Vel), and flight time.
In particular, we adopt flight time as the primary efficiency metric because it inherently reflects both distance and velocity performance.

\begin{table}[t]
    \centering
    \caption{Average performance of planners on random maps.}
    \label{tab:random_map_performance}
    \setlength{\tabcolsep}{3pt}
    \small
    \begin{tabular}{@{}lcccc@{}}   
        \toprule
        \textbf{Method}
        & \shortstack{Flight Dist\\(m)}
        & \shortstack{Avg Vel\\(m/s)}
        & \shortstack{\textbf{Flight Time}\\(s)}
        & \shortstack{\textbf{Safety}\\(Obs Count)} \\  
        \midrule
        EGO-Planner                            & 28.72 & 0.68 & 42.13 & 3.2 \\
        ViGO                                   & 29.26 & 0.63 & 45.96 & 2.5 \\
        Medial-based                           & 30.33 & 0.41 & 73.63 & \textbf{0.5} \\
        Safe EGO                               & 28.61 & 0.69 & 41.36 & 2.2 \\
        \shortstack[l]{\textbf{Ours}\\[-2pt]\textbf{(RTG = 0.5)}}  & 32.35 & 0.59 & 54.54 & \underline{\textbf{1}} \\
        \shortstack[l]{\textbf{Ours}\\[-2pt]\textbf{(RTG = 10.0)}} & 29.47 & 0.74 & \underline{\textbf{39.47}} & 2 \\
        \bottomrule
    \end{tabular}
\end{table}

Experimental results are summarized in Tab. \ref{tab:grid_performance}, which shows averages from five experiments conducted on grid maps with different obstacle spacings.
The experimental results indicate that our planner achieves the best efficiency with higher RTG settings compared to the baseline planners. 
Moreover, when tuned for safety (with lower RTG settings), our planner produces the lowest DTW error, demonstrating superior safety performance.
This confirms that our proposed planner can adjust the safety–efficiency trade-off using only the RTG within a single trained model, while also outperforming the baseline planners.

In contrast, the Safe EGO-Planner (Safe EGO), which doubles the safety weight of the standard EGO-Planner, exhibits performance nearly identical to that of the standard version.
This demonstrates that although conventional planner include safety parameters, effectively adjusting the safety–efficiency trade-off using these parameters remains highly challenging.
Fig. \ref{fig:grid_map_comparison} shows trajectories generated by our planner in grid maps with obstacle spacings of 4 m and 5 m (Grid map 4, Grid map 5) under different RTG settings.
It also presents the resulting DTW error.

\begin{figure}[t!]
    \centering
    \includegraphics[width=0.7\linewidth]{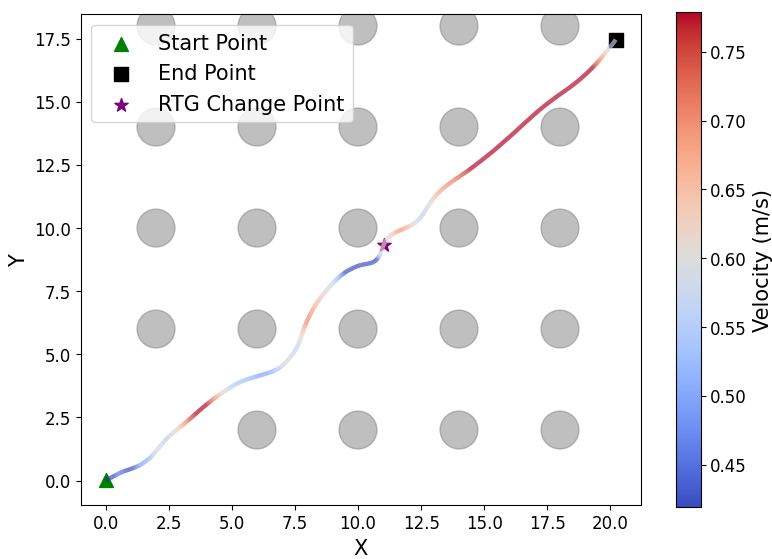}
    \caption{Trajectory under RTG change in real-time.}
    \label{fig:dynamic_rtg_change}
\end{figure}

\begin{figure}[t!]
    \centering
    \includegraphics[width=0.65\linewidth]{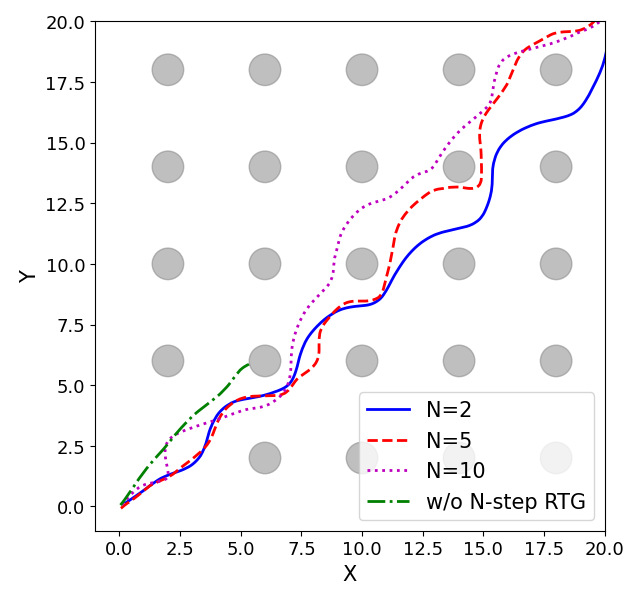}
    \caption{Trajectories with varying $N$ in $N$-step RTG.}
    \label{fig:ablation_study}
\end{figure}

\subsubsection{Random Map Experiments}
In random map experiments, safer trajectories than those generated by the medial axis-based planner can exist.
Therefore, we quantify safety by counting the number of obstacles within 0.5 m (Obs Count) during flight rather than using DTW error.
The efficiency metric is applied exactly as in the grid map environment.
The random maps, which include environments unseen during training, were evaluated over five experiments, and the average performance is summarized in Tab. \ref{tab:random_map_performance}. 
Similar to the grid map experiments, our approach shows that simply tuning the RTG allows our planner to adjust safety and efficiency dynamically in complex, unstructured environments. 
Moreover, the random map results also reveal that higher RTG settings yield the best efficiency.
In contrast, lower RTG settings achieve the most favorable safety metrics among all methods except for the medial axis-based planner.
Fig. \ref{fig:random_map_comparison} illustrates how different RTG settings lead to distinct trajectory behaviors in random maps.

\subsubsection{Adjusting RTG during Trajectory Generation}
We further evaluated the ability of our proposed algorithm to adjust the safety–efficiency trade-off dynamically by changing the RTG during trajectory generation.
As shown in Fig. \ref{fig:dynamic_rtg_change}, our planner operates at a low RTG setting before the RTG change point, generating a safe trajectory.
When the RTG is increased at the change point, the planner produces a more efficient trajectory, and the drone's velocity rises accordingly.
These results confirm that our planner can dynamically modulate safety and efficiency during flight simply by adjusting the RTG in real-time.

\subsubsection{Ablation Study}
We conducted an ablation study on the effectiveness of the proposed $N$-step RTG.
The experiments were performed on Grid map 4, comparing the performance of our planner with a fixed RTG of 0.5 under various conditions: without $N$-step RTG (w/o $N$-step RTG), $N=10$, $N=5$, and our chosen experimental setting, $N=2$.
As shown in Fig. \ref{fig:ablation_study}, the model trained without $N$-step RTG failed to learn a stable trajectory, leading to collisions with obstacles.
When $N=10$, the trajectory exhibited unclear trade-off behavior, initially following a safe route but eventually becoming unreasonable.
For $N=5$, the trajectory maintained a safe route longer than the $N=10$ case but eventually became unstable.
Finally, our setting ($N=2$) produced stable and consistent trajectories, validating that the proposed $N$-step RTG effectively improves the model's ability to estimate and maintain the desired safety–efficiency trade-off.

\subsection{Real-World Experiments}

\begin{figure}[!t]
    \centering
    \includegraphics[width=0.4\textwidth]{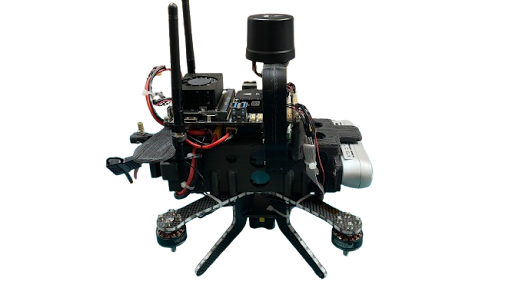}
        
    \begin{minipage}{0.5\textwidth}
        \centering
        \footnotesize
        \begin{tabular}{c|c}
            \toprule
            Hardware Components of Drone & Model \\
            \midrule
            Flight Controller & Pixhawk v6 \\
            Mission Computer & Jetson Orin Nano 8GB \\
            Stereo Camera & Intel RealSense D455 \\\
            RTK-GPS & F9P-Ultralight \\
            \bottomrule
        \end{tabular}
        \label{tab:performance}
    \end{minipage}
    
    \caption{The drone used in real-world experiments.}
    \label{fig:drone}
\end{figure}

\begin{figure}[t!]
    \centering
        \begin{subfigure}{0.43\textwidth}
        \centering
        \includegraphics[width=\linewidth]{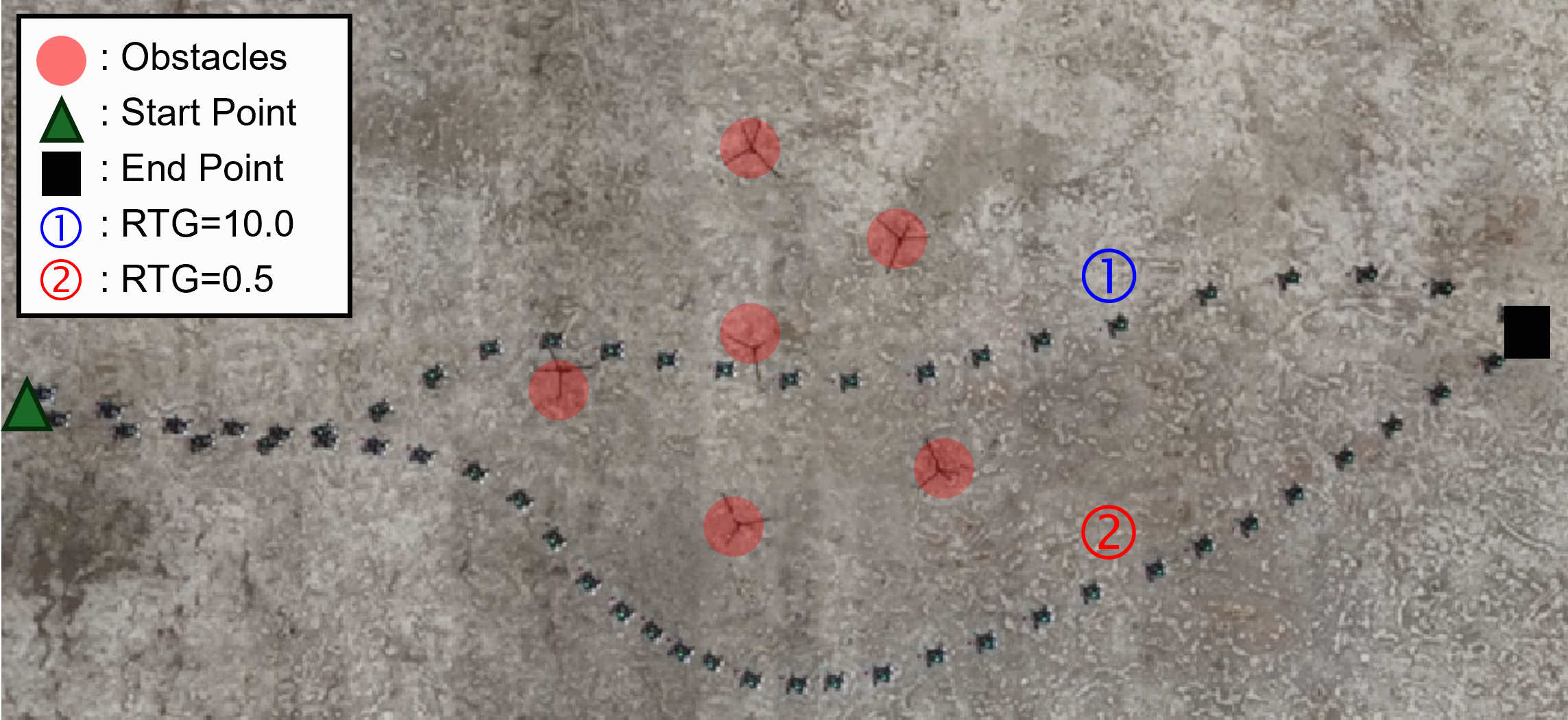}
        \caption{Stacked top-view image in the real-world experiments.}
        \label{fig:real_traj}
    \end{subfigure}
    \begin{subfigure}{0.41\textwidth}
        \vspace{0.02\linewidth}
        \centering
        \includegraphics[width=\linewidth]{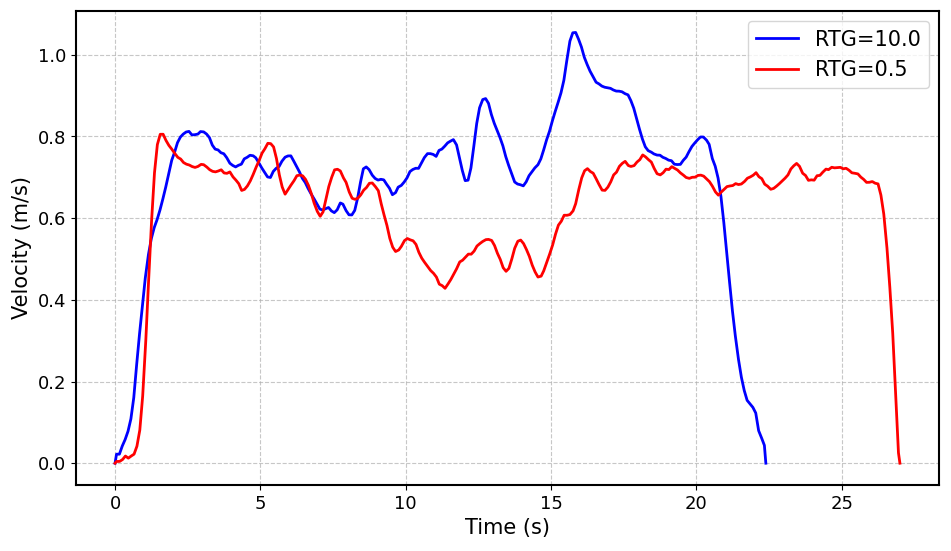}
        \caption{Comparison of velocity in the real-world experiments.}
        \label{fig:real_traj_plot}
    \end{subfigure}
    \caption{Results of real-world experiments.}
    \label{fig:real_experiments}
\end{figure}

We conducted real-world experiments to validate the practicality of our proposed algorithm.
Fig. \ref{fig:drone} shows the drone used in our real-world experiments.
In these experiments, obstacles were densely arranged along the direct path between the start and end points.
Fig. \ref{fig:real_traj} shows stacked top-view images comparing trajectories with high and low RTG settings.
Fig. \ref{fig:real_traj_plot} compares flight velocities under these two RTG settings.
Consistent with our simulation results, a higher RTG setting produced a direct and efficient trajectory through the obstacle cluster, while a lower RTG resulted in a safer trajectory avoiding obstacles.
Additionally, the drone exhibited lower average velocity under the lower RTG setting, aligning with the simulation findings.
These results confirm that our planner can dynamically adjust the safety–efficiency trade-off not only in simulation but also under real-world conditions.

\section{CONCLUSIONS}

We propose a DT–based planner that uses RTG as a temperature parameter to dynamically adjust the trade‑off between the safety and efficiency of trajectories.
Our key innovation is that a single trained DT model can exhibit a wide range of behaviors simply by tuning the RTG input.
This contrasts sharply with traditional methods that require expert knowledge for parameter tuning, separate algorithms, or retraining to achieve different trade-offs.
It is particularly noteworthy that our DT-based planner was trained exclusively using offline data collected from various state-of-the-art planners, including our own medial axis-based planner designed explicitly to account for safety margins.
Our experiments in simulation environments demonstrate that our proposed algorithm can adapt to various mission requirements while outperforming baseline planning algorithms.
Moreover, real-world experiments confirm that our approach maintains its adaptability and robustness outside simulation.
In future work, we plan to extend the approach to multi-agent scenarios and explore meta-planning strategies to automatically determine optimal RTG values based on environmental context.

\addtolength{\textheight}{-1cm}   




\section*{ACKNOWLEDGMENT}

This work was partly supported by Innovative Human Resource Development for Local Intellectualization program through the Institute of Information \& Communications Technology Planning \& Evaluation (IITP) grant funded by the Korea government (MSIT) (IITP-2025-RS-2020-II201462), the Unmanned Vehicle Advanced Technology Development Program, grant funded by the Ministry of Science and ICT (2023M3C1C1A01098416), and the Basic Science Research Program through the National Research Foundation of Korea (NRF) funded by the Ministry of Education (2018R1A6A1A03025526).



\end{document}